\documentclass[conference]{IEEEtran}
\IEEEoverridecommandlockouts
\usepackage{cite}
\usepackage{amsmath,amssymb,amsfonts}
\usepackage{algorithmic}
\usepackage{graphicx}
\usepackage{textcomp}
\usepackage{xcolor}

\usepackage{hyperref}       
\usepackage{url}            
\usepackage{booktabs}       
\usepackage{nicefrac}       
\usepackage{microtype}      
\usepackage{lipsum}
\usepackage{fancyhdr}       
\usepackage{verbatim}

\usepackage{etoolbox}
\usepackage{times}
\usepackage{epsfig}
\usepackage{adjustbox}
\usepackage{multicol}
\usepackage{siunitx}
\usepackage{blindtext}
\usepackage{caption}
\usepackage{balance}
\usepackage{subcaption}
\def\BibTeX{{\rm B\kern-.05em{\sc i\kern-.025em b}\kern-.08em
    T\kern-.1667em\lower.7ex\hbox{E}\kern-.125emX}}

\makeatletter
\def\@IEEEpubidpullup{8\baselineskip}
\patchcmd{\@IEEEkeywords}{\hfil\IEEEkeywordsname}{\hfil Keywords}{}{}
\makeatother

\begin{document}

\title{FishNet: Deep Neural Networks for Low-Cost Fish Stock Estimation\\
}

\author{\IEEEauthorblockN{Moseli Mots'oehli}
\IEEEauthorblockA{\textit{Information and Computer Sciences} \\
\textit{University of Hawai'i at Manoa}\\
moselim@hawaii.edu}
\and
\IEEEauthorblockN{Anton Nikolaev}
\IEEEauthorblockA{\textit{Information and Computer Sciences} \\
\textit{University of Hawai'i Manoa}\\
nikolaev@hawaii.edu}
\and
\IEEEauthorblockN{Wawan B. IGede}
\IEEEauthorblockA{
\textit{Yayasan Konservasi Alam Nusantara} \\
\textit{People and Nature Consulting International}\\
gede.wawan@gmail.com}
\and
\IEEEauthorblockN{John Lynham}
\IEEEauthorblockA{\textit{Economics} \\
\textit{University of Hawai'i Manoa}\\
lynham@hawaii.edu}
\and
\IEEEauthorblockN{Peter J. Mous}
\IEEEauthorblockA{
\textit{Yayasan Konservasi Alam Nusantara}\\
\textit{People and Nature Consulting International}\\
pmous@ykan.or.id}
\and
\IEEEauthorblockN{Peter Sadowski}
\IEEEauthorblockA{\textit{Information and Computer Sciences} \\
\textit{University of Hawai'i Manoa}\\
peter.sadowski@hawaii.edu}
}

\maketitle

\begin{abstract}
Fish stock assessment often involves manual fish counting by taxonomy specialists, which is both time-consuming and costly. We propose FishNet, an automated computer vision system for both taxonomic classification and fish size estimation from images captured with a low-cost digital camera. The system first performs object detection and segmentation using a Mask R-CNN to identify individual fish from images containing multiple fish, possibly consisting of different species. Then each fish species is classified and the length is predicted using separate machine learning models. To develop the model, we use a dataset of 300,000 hand-labeled images containing 1.2M fish of 163 different species and ranging in length from 10~cm to 250~cm, with additional annotations and quality control methods used to curate high-quality training data.  On held-out test data sets, our system achieves a 92\% intersection over union on the fish segmentation task, a 89\% top-1 classification accuracy on single fish species classification, and a 2.3~cm mean absolute error on the fish length estimation task.
\end{abstract}


\section{Introduction}
Predictions that all of the world’s stocks of commercially important fish could collapse by the year 2048~\cite{worm2006impacts} have been tempered by recent evidence that fish stocks are recovering in many high-income countries~\cite{worm2009rebuilding}. This turnaround is typically attributed to stringent catch limits backed by an accurate scientific assessment of the number of fish in a particular stock (population). It is challenging to set a limit on catch appropriately unless scientists know how many fish are actually present in a wild population. Despite progress in many high-income countries, the majority of the world’s fisheries remain “unassessed” and the prognosis for these fisheries is concerning: populations are declining and are on a trajectory towards functional extinction~\cite{costello2012status}. One of the major barriers to performing fish stock assessment in the developing world is the cost. The federal government in the US spends approximately $\$215$ million a year on fish stock assessment \cite{Merrick:NOAA16}, which excludes spending by state governments on assessment of stocks within three nautical miles of the coast (state waters). As an example, the average cost of a fish stock assessment performed by NOAA’s Pacific Islands Fisheries Science Center in Honolulu is $\$5.6$ million, which exceeds the total value of many developing country fisheries.

\begin{figure}[!tbp]
    \includegraphics[width=\linewidth]{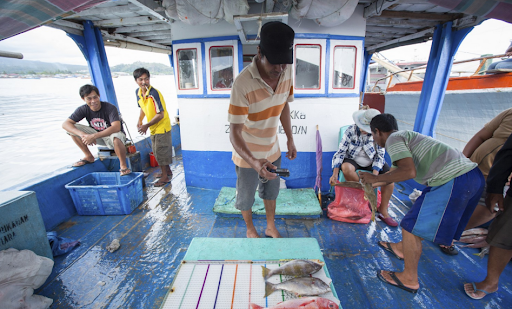}
    \caption{Fishers in Indonesia on their boats photographing their catch on a standard color-coded measuring board. Source: Ed Wray, for Yayasan Konservasi Alam Nusantara.}
    \label{fig:fishers}
\end{figure}

Costly stock assessment also prevents many well-managed fisheries from accessing lucrative markets for sustainable seafood. For example, the Marine Stewardship Council (MSC), which endorses seafood with its blue eco-label requires a detailed stock assessment of the fishery for certification. In \cite{perez2016marine}, the authors argue that the high costs of certification remain a barrier to many fisheries in Latin America and the Caribbean accessing the benefits of seafood ecolabels. For example, there is only one fishery certified as sustainable in Indonesia by the MSC, and this is, in part, because the species being targeted (yellowfin tuna) occurs throughout the Pacific Ocean; the expensive assessment of this stock is performed by an intergovernmental body, funded by the US, Australia, France, Japan, and a number of other countries. Here we propose a methodology for drastically reducing the cost of fisheries stock assessment by combining citizen science with machine learning.

Advances in digital photography, computer vision, and artificial intelligence (AI) make automating fish-stock estimation an attractive alternative. To address this challenge, we propose FishNet, a novel computer vision system for automated fish species classification and size estimation in a tropical snapper-grouper fishery in Indonesia. This is one of the most difficult fisheries in the world in which to perform a stock assessment, with small-scale fishers catching over a hundred different species, using a variety of fishing gear, spread across Indonesia’s 17,000 different islands. As part of a participatory science program, The Nature Conservancy in Indonesia gave small-scale fishers digital cameras and asked them to photograph all their catch against a consistent background with color-coded measurement scales, as seen in Figure~\ref{fig:fishers}. A team of marine biologists then identified the species and length of fish from the photographs, resulting in 300,000 images containing approximately 1.2 million fish annotated with species and length. Because the system requires no special equipment beyond a digital camera and a standardized board with fiduciary markings, the approach is a feasible and financially practical solution to the problem of fish stock estimation at scale.

\section{Related Work}\label{Literature}

\textbf{Taxonomic Classification:}
Modern computer vision systems such as Facebook's Detectron2 \cite{wu2019detectron2} use deep neural network models to perform object detection, segmentation, and classification. These models are pre-trained on the ImageNet-1k dataset and can be fine-tuned to application-specific tasks through transfer learning. Previous works have used this approach to classify fish for fish stock estimation~\cite{Prasetyo:VGGNetFishClass,Dhruv:underWaterClass17,Guang:AutoFishClass17,Oguzhan:LargeScaleFishSegmentation20}, relying on specialized and expensive cameras. In \cite{Prasetyo:VGGNetFishClass,qin2016deepfish}, the authors use a pre-trained convolutional neural network (CNN) architecture and the Fish4Knowledge Ground-Truth dataset \cite{Jessica:fish4knowledgeDataset12} for fish species classification. However, this dataset contains a much smaller number of images (27,370) of underwater fish extracted from video recordings and only 23 species, in comparison to the 163 species in this work. Kandimalla et al. present an automated framework for detecting, classifying, and tracking fish using high-resolution imaging sonar and underwater videos~\cite{Kandimalla:AutoFishDetectioClassification22}. While it shares similarities with our work in addressing the classification and counting of multiple fish, it focuses solely on eight fish species and does not estimate the size of each fish. A similar approach is used in \cite{Connolly:improvedbaited2021} where underwater video is used for detection and fish count estimation. This system requires significantly more cost as underwater images require specialized cameras and hardware, and processing video requires significantly more computation.
The work in \cite{Garcia:DeepFishAllIn22} is similar to ours as they perform detection, multi-fish segmentation, classification, and size estimation. However, they use a smaller dataset (DeepFish) of 1,291 labeled images with 7,339 fish of 59 species, compared to our 1.2M images of more than 160 species. They also focus on the retail side of the problem where measurements and labeling are derived from fish markets and the fish are market-sized as opposed to all fish caught at sea (which is more relevant for fish stock assessment).
The work of \cite{Garcia:DeepFishAllIn22} focuses solely on multi-fish images where all fish are assumed to be of the same species. In contrast, our system is more versatile, and is capable of identifying and estimating the size of each fish in images containing multiple species. Furthermore, their focus is on large fish like bluefin tuna and sharks, while our system is trained to identify and measure fish ranging from 10cm to over 200cm in size. To our knowledge, our work offers the most comprehensive and practical approach to fish stock estimation, covering segmentation, classification, and size estimation using low-cost camera images on a dataset of over 160 species.

\textbf{Fish Size Estimation:} To help estimate fish size from images, our system uses a fiduciary marker of known-length: a 10cm rectangular colored areas on the sides of each measuring board. These fiduciary markers help estimate the size of the fish even when the photos are taken from different distances and angles. In \cite{Alvarez:SizeEstimation}, a mask R-CNN similar to the one used in this work detects and measures the length of fish heads and bodies, but only for a single species. In \cite{Monkman:RegionalFishLength19}, the authors place three \href{https://pyimagesearch.com/2020/12/21/detecting-aruco-markers-with-opencv-and-python/}{ArUco} fiduciary markers of different sizes on polypropylene sheets, along with the fish to be measured. The authors extend their work in \cite{Monkman:photogrammety20}, by using the same three fiduciary markers and estimating fish length using standard smartphone cameras. They use mask R-CNN to detect and segment objects of interest. Their evaluation uses a smaller dataset of 1,000 images with fewer fish species and doesn't include species classification as all fish are assumed to be the same species. This limits its applicability for general automated fish stock estimation. In contrast, FishNet is more general, accurately estimating sizes for a broader range of 163 fish species ranging from 10cm to over 200cm.

\section{Methods}\label{methods}

\begin{figure}[!tbp]
    \includegraphics[width=\linewidth]{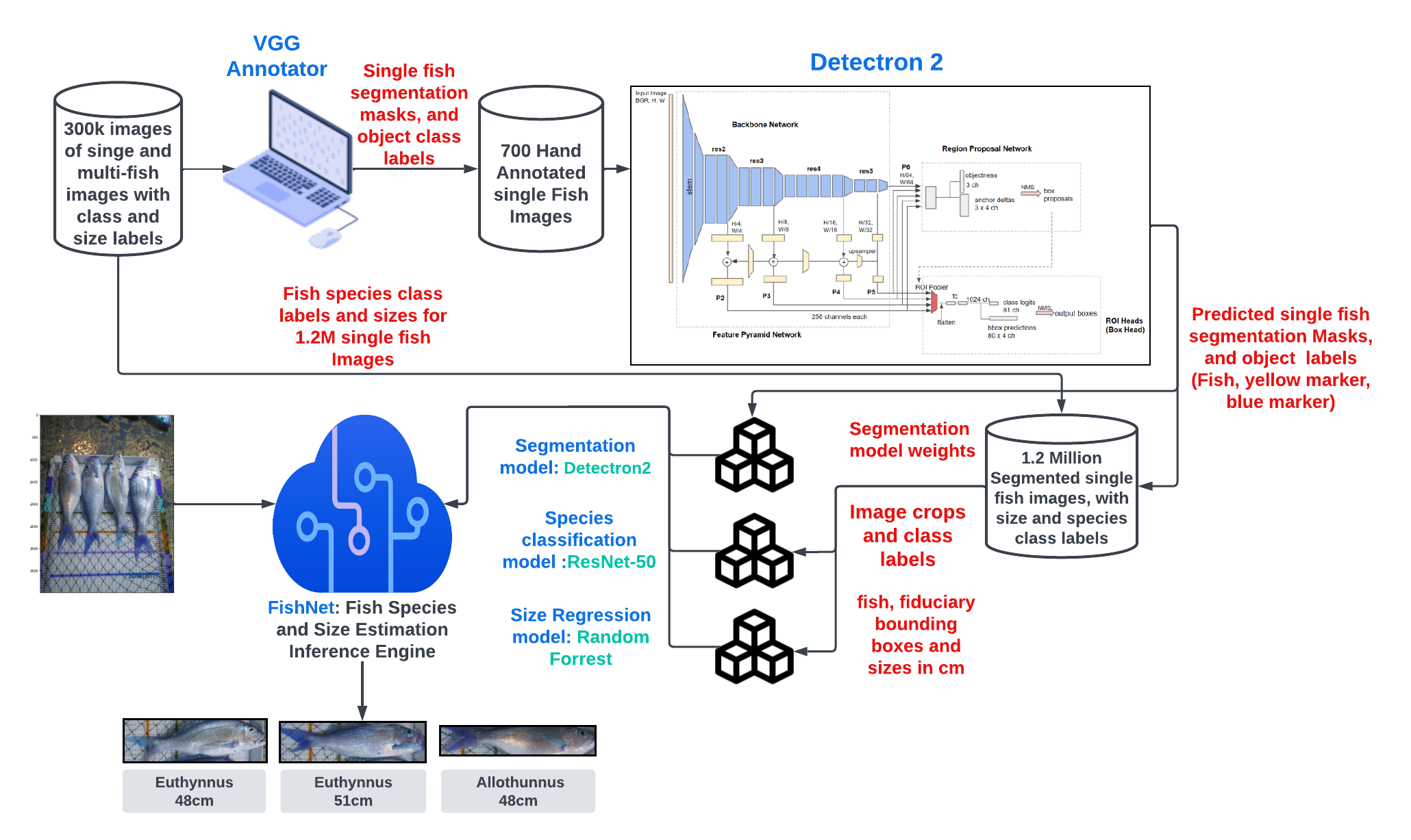}
    \caption{FishNet's training process, starting from data labeling to achieving high-accuracy fish classification and size estimation. It shows the data flow between datasets and models.}
\label{fig:fishnet}
\end{figure}

In this section, we outline the implementation of data collection, image segmentation, classification, and fish length regression. Figure~\ref{fig:fishnet} shows the complete FishNet system: the models, outputs, and training datasets.

\subsection{Data Collection}\label{methods_data_collection}
CODRS (Crew Operated Data Recording Systems) is a program run
by The Nature Conservancy's Indonesia office. Data on species and size distributions of catch (as needed for length-based stock assessments) are collected via photographs taken on digital cameras by crew onboard fishing vessels. The fish are positioned over standardized measuring boards (1-by-0.8~m) before the photographs are taken so their length can be easily inferred (see Figure~\ref{fig:labeled_fish}). The whiteboard, with multi-colored markings every 10~cm serves as a fiduciary marker for scale and perspective. The images often contain multiple fish and photographs are taken from a variety of distances, angles, and lighting conditions. At the end of each month, the memory card from each camera is handed over for the processing of the images by fish identification experts working for The Nature Conservancy. Processing includes identification of the species and length of the fish, double-checking by a data quality control coordinator, and storage in an online database known as i-Fish. Participation in the program is voluntary and fishers receive a stipend for participating. Over 300 vessels have cooperated and contributed to the database. The set of 300,000 photographs was then hand-annotated by taxonomic experts to label each fish with species and length. Approximately 1.2 million fish were hand-labeled in this manner, with 163 different species represented. 

\begin{figure}[!tbp]
    \includegraphics[width=\linewidth]{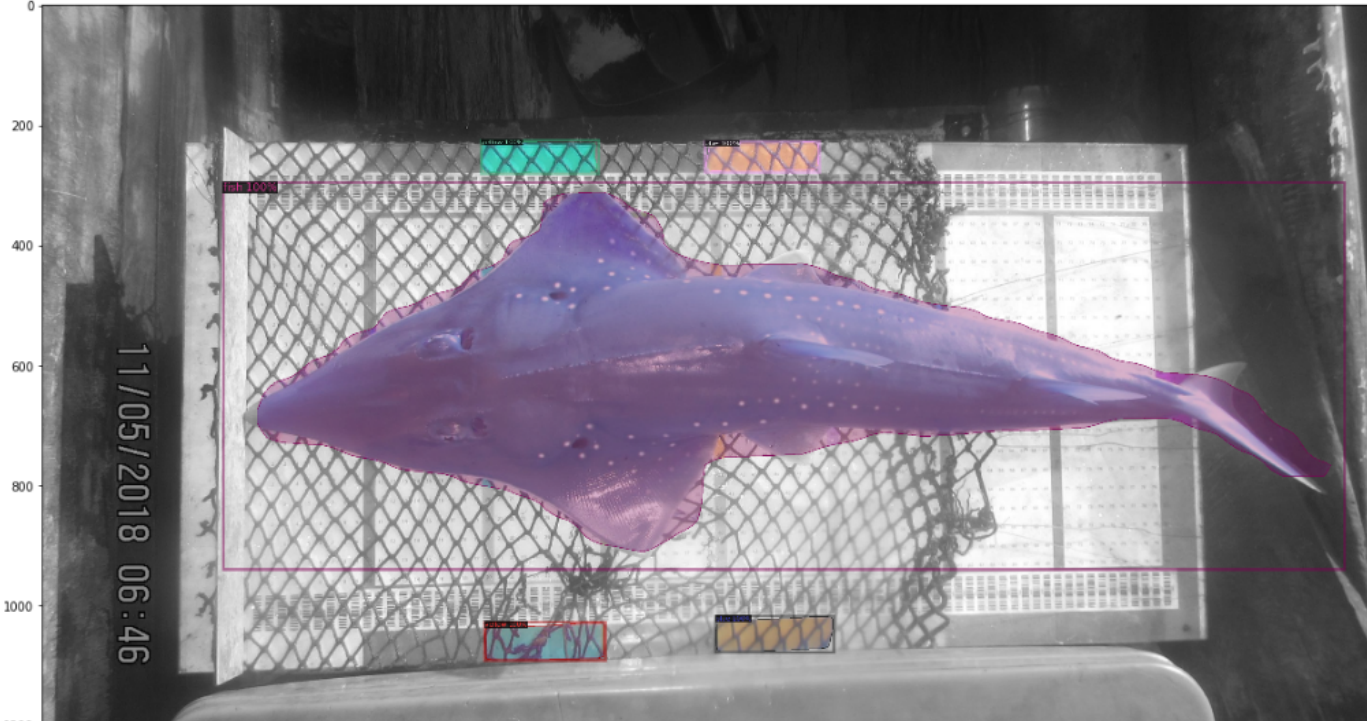}
    \caption{Example of an image segmented using Detectron2 to identify both the fish and the 10~cm colored rectangles used as fiduciary markers. The original image is in color, but has been shown in grayscale to show the segmentation. Source: Yayasan Konservasi Alam Nusantara.}
    \label{fig:fish_segmentation}
\end{figure}

\subsection{Data Curation}\label{methods_data_curation}

Large-scale data-labeling projects nearly always contain some errors. In this work, one source of error is that photographs containing multiple fish are annotated with a list of species and fish lengths, but the order of the list does not always match the order of the fish in the image. In most images, fish are labeled from top-to-bottom, left to right, but many examples were found in which this is not the case, leading to ambiguity in the annotations. Therefore, the species annotations for images containing a single fish were assumed to be reliable, while the species annotations for images containing multiple fish are assumed to be noisy. 

In experiments, we used the most reliable data for training, but there is interest in developing methods for curating the rest of the data. We developed an algorithm that attempts to disambiguate the species annotations using the species classification prediction. For images with one or two fish, matching the labels to the fish is trivial as there are only one or two possible combinations. For images with three fish and above on a board, we use the output of a species classification to assign a likelihood for each object permutation, and selected the one with the highest likelihood. 

\subsection{Image Segmentation}\label{methods_segmentation}
The FishNet system first applies an object detection and segmentation model to predict a bounding box and segmentation mask around each individual fish and fiduciary color box. For this, we use Metas's Detectron2~\cite{wu2019detectron2} implementation of Mask R-CNN~\cite{maskrcnn2017} with a Resnet-50 backbone~\cite{he2016deep}. This model has been pre-trained on the Microsoft Common Objects in Context (MS COCO) dataset~\cite{Lin:MSCOCO14} and fine-tuned on a small self-annotated dataset. 
    
\begin{figure}
    \centering
    \includegraphics[width=\linewidth]{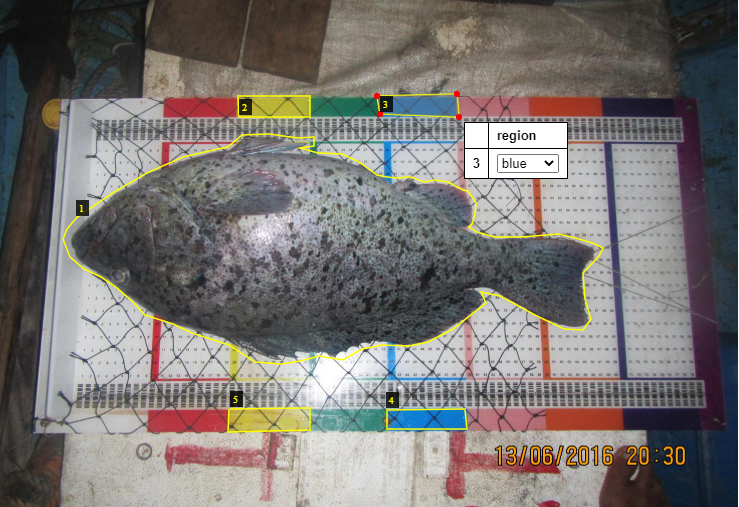}
    \caption{Image of a fish along with the yellow and blue reference color boxes on either side of the boards. In this example, a human annotator has outlined these objects using the VGG Annotation tool. Source: Yayasan Konservasi Alam Nusantara.}
    \label{fig:labeled_fish}
\end{figure}

The segmentation model is fine-tuned on a set of random images that were manually annotated for object detection and segmentation. Annotators segmented each fish and a set of fiduciary markers consisting of four 3-by-10 cm colored boxes (two yellow and two blue) located on the edges of the presentation board. The Visual Geometry Group (VGG) Image Annotation (VIA) tool~\cite{dutta2019vgg} was used to annotate 700 random images containing multiple fish of the same species with polygon outlines of three object types: (1) fish, (2) yellow boxes, and (3) blue boxes. When these objects are partially occluded (e.g., by humans, shadows, or fishing nets), the annotator infers their shape as shown in Figure~\ref{fig:labeled_fish}.  Of the 700 images, 420 (60\%) were used for training, 140 (20\%) were used for cross-validation, and 140 (20\%) used for testing.

Given a new image, the fine-tuned Detectron2 model detects objects of interest with (1) a bounding box enclosing the object, (2) a pixel-wise segmentation mask, and (3) the predicted class (fish, yellow box, or blue box).
During inference, FishNet uses the segmentation mask from Detectron2 to crop individual fish images for input to classification and size regression models. We evaluated the segmentation model in terms of accuracy and the Intersection over Union (IoU) score on the held-out validation set. These metrics inform us of how often we are detecting and correctly assigning objects, as well as how well the segmentation mask covers the object since these outputs directly feed into the length estimation algorithm. A high IoU score (near 100\%) means the bounding boxes fully and tightly enclose the fish, and this will result in better data for the downstream classification and regression tasks.

\subsection{Species Classification}\label{methods_classification}
For species classification, a pre-trained ResNet-50 model was fine-tuned on a dataset of 50,000 images containing a single fish, cropped using the bounding box supplied by the object detection model. We use 30,000 images (60\%) for training, 10,000 images (20\%) for validation, and 10,000 images (20\%) as a test set. For training, the learning rate was $0.005$, which was reduced linearly by 10\% when the validation set error does not decrease for 10 epochs. Various augmentation strategies are applied during training to improve generalization, as listed in Table~\ref{tab:classification_transformations}. During testing,  the images are re-scaled to $152 \times 152$ and normalized.

\begin{table}
\begin{center}
\begin{tabular}{l|c}
\hline
Transformation & Range/Value \\
\hline\hline
Intensity normalization & [0,1]\\
Random Rotation & [0, \ang{30}] \\
Width, length, channel shift, and random zoom & [0, 0.2] \\
Shear & [0, 0.3]\\
Feature-wise normalization and horizontal flip & Yes\\
Fill mode & nearest\\
Re-scale & $152\times152$\\
\hline
\end{tabular}
\end{center}
\caption{Data augmentations applied during training of the species classification model.}
\label{tab:classification_transformations}
\end{table}

\subsection{Fish Length Estimation}\label{methods_regression}
Length regression is a two-step process. First, we use Detectron2 to detect and locate fish and colored fiduciary markers, which provide scale information due to their fixed and known size. Images without a visible fiduciary marker are discarded from analysis. A random forest regression model is trained to predict fish size in centimeters from a set of features that are extracted from the image and the detected fish and fiduciary markers:

\begin{enumerate}
    \item {Color Boxes:} Color box count, median color box length, mean color box length, median color box segment length, and maximum color box segment length. 
    \item {Detected Fish:} The confidence score, bounding box length, and segment mask length for each fish. We use both measurements since the segment mask can be slightly shorter than the actual fish due to less than $100\%$ IoU in fish segmentation.
\end{enumerate}

\noindent
The features include redundant information which helps improve predictions under occlusion or poor lighting conditions. We chose to use a random forest due to its robustness to outliers.

\section{Results}\label{results}

In experiments, we evaluate each component of the FishNet model: detection, segmentation, species classification, and length regression. Due to the limitations of human annotations discussed above, we report results on held-out subsets of the full dataset that have high-quality annotations.

\subsection{Detection}

\begin{figure}
\centering
\includegraphics[width=\linewidth]{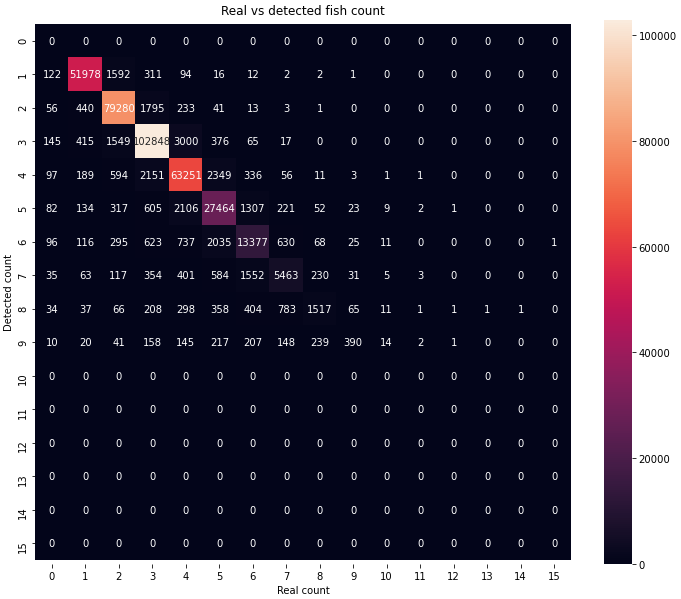}
\caption{Heat map of detected vs. ground truth fish count in multi-fish images. Diagonal entries indicate where the model correctly detects and segments the number of fish}
\label{fig:count_heatmap}
\end{figure}

\begin{figure}
\centering
\includegraphics[width=\linewidth]{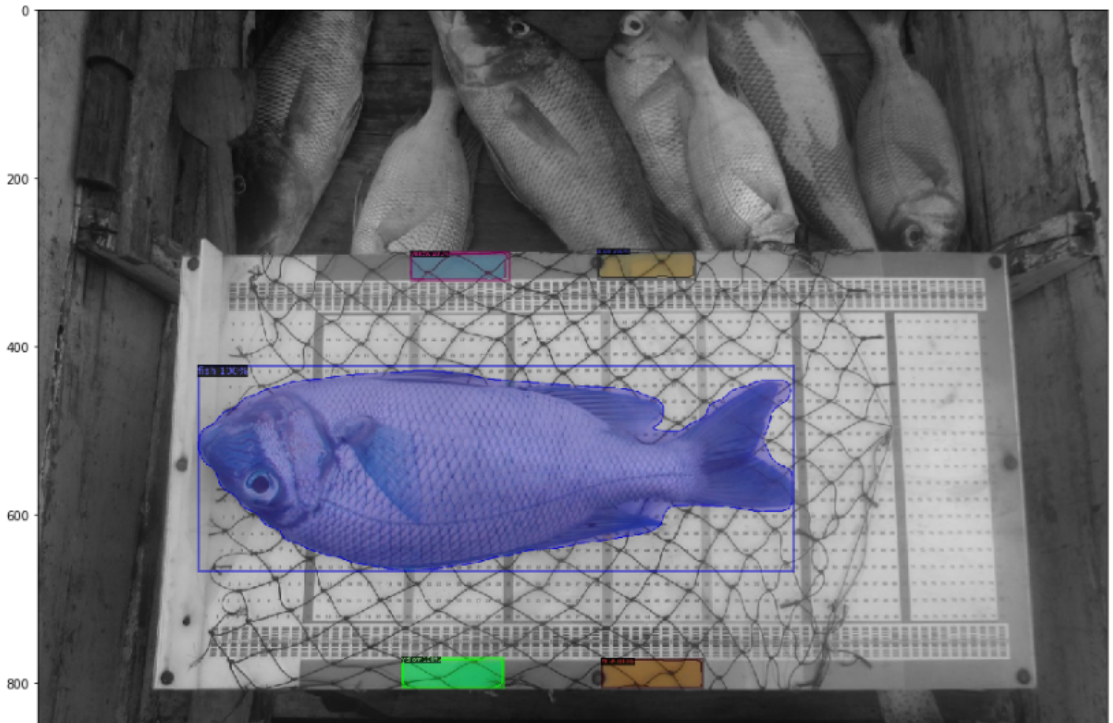}
\caption{Image containing multiple fish with the model successfully detecting only fish placed on the board. Source: Yayasan Konservasi Alam Nusantara.}
\label{fig:background_fish}
\end{figure}

The performance of the object detection model was evaluated by comparing the number of fish detected in each image with the number of fish listed by human annotators. On a random subset of 10,000 images containing only a single fish, the model detects fish in 99\% of the images and fiduciary markers in 97\% of the images. For the more challenging case where an image can contain multiple fish, we ran the object detection model on all 300,000 images and observed that the predicted number of fish exactly agreed with the human annotator in 92\% of images. Figure~\ref{fig:count_heatmap} shows the confusion matrix for detected vs. human-annotated number of fish per image. When the detection model is wrong, it is usually off by only one or two fish, overestimating the count in 3\% of images and underestimates the count in 5\% of images. The model is more likely to miscount fish when there are more on the board, especially smaller fish, due to overlap. Some of these errors are due to fish located in the background of the image, as shown in Figure~\ref{fig:background_fish}, and could be eliminated by more careful image-capture procedures or by cropping the images as a pre-processing step.

\subsection{Segmentation}

\begin{figure}
\centering
\includegraphics[width=\linewidth]{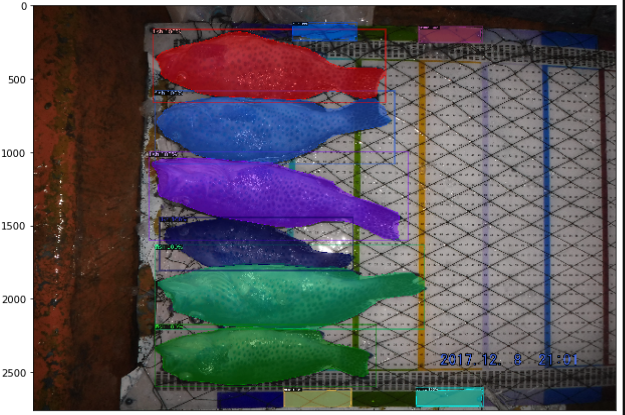}
\caption{Detectron2 detects six fish and four fiduciary markers located at the edges of the presentation board.}
\label{fig:good_prediction_example}
\end{figure}

\begin{figure}
\centering
\includegraphics[width=\linewidth]{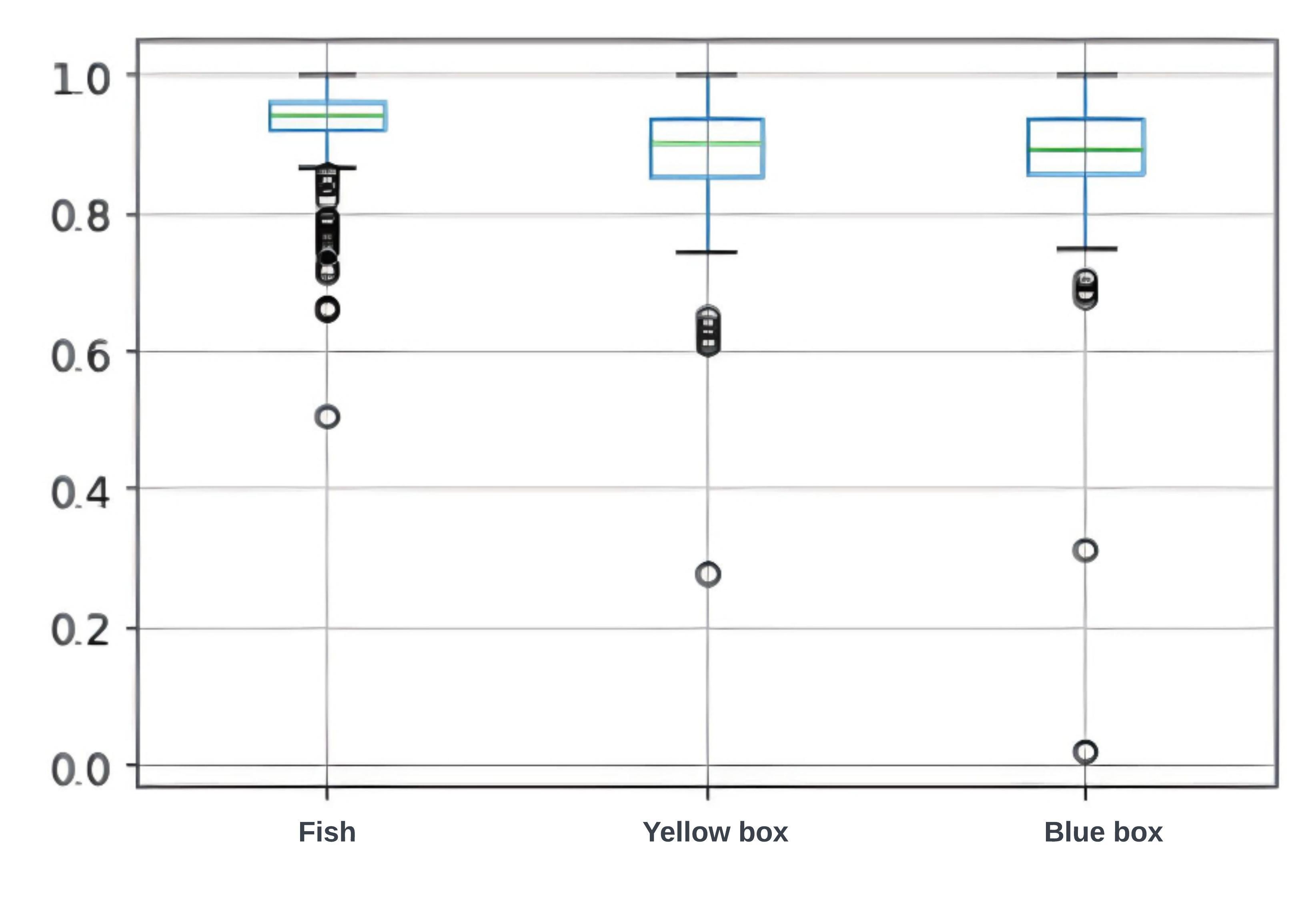}
\caption{Performance of the segmentation model in terms of IoU on the validation set. The box and whisker plot show the minimum, first quartile, median, third quartile, and maximum values. Circular dots represent outliers, where a fish is missed entirely or partially segmented due to occlusion.}
\label{fig:IoU_results}
\end{figure}

Segmentation performance was evaluated using a held-out test set of 140 (20\% of 700) images containing multiple fish and segmented by human annotators. The predicted segmentation masks have a mean IoU of 92\% (median 94\%) for fish and a mean IoU of 86\% (median 88\%) for fiduciary markers, as shown in the box and whisker plot in Figure~\ref{fig:IoU_results}. Figure ~\ref{fig:good_prediction_example} is an example of predictions on an image where a high IoU is achieved for all the fiduciary markers and fish on the board. The performance of the detection and segmentation model affects species classification and length regression. A high IoU score indicates bounding boxes tightly enclose the fish, leading to better training data for length estimation in single fish images. In multi-fish images, poor IoU from Detectron2 affects both classification and regression. Loose bounding boxes can overestimate lengths and cause parts of one fish to appear in the crop of another, making it harder to learn critical features that explain the differences in species.

\subsection{Species Classification}\label{results_classification}
Species classification using the fine-tuned ResNet-50 model was evaluated using the held out test set of 10,000 single-fish images. The model achieves 89\% top-1 classification accuracy and $97\%$ top-5 accuracy on the test set.

\subsection{Length Regression}\label{results_single_fish_regression}

\begin{figure}
\includegraphics[width=\linewidth]{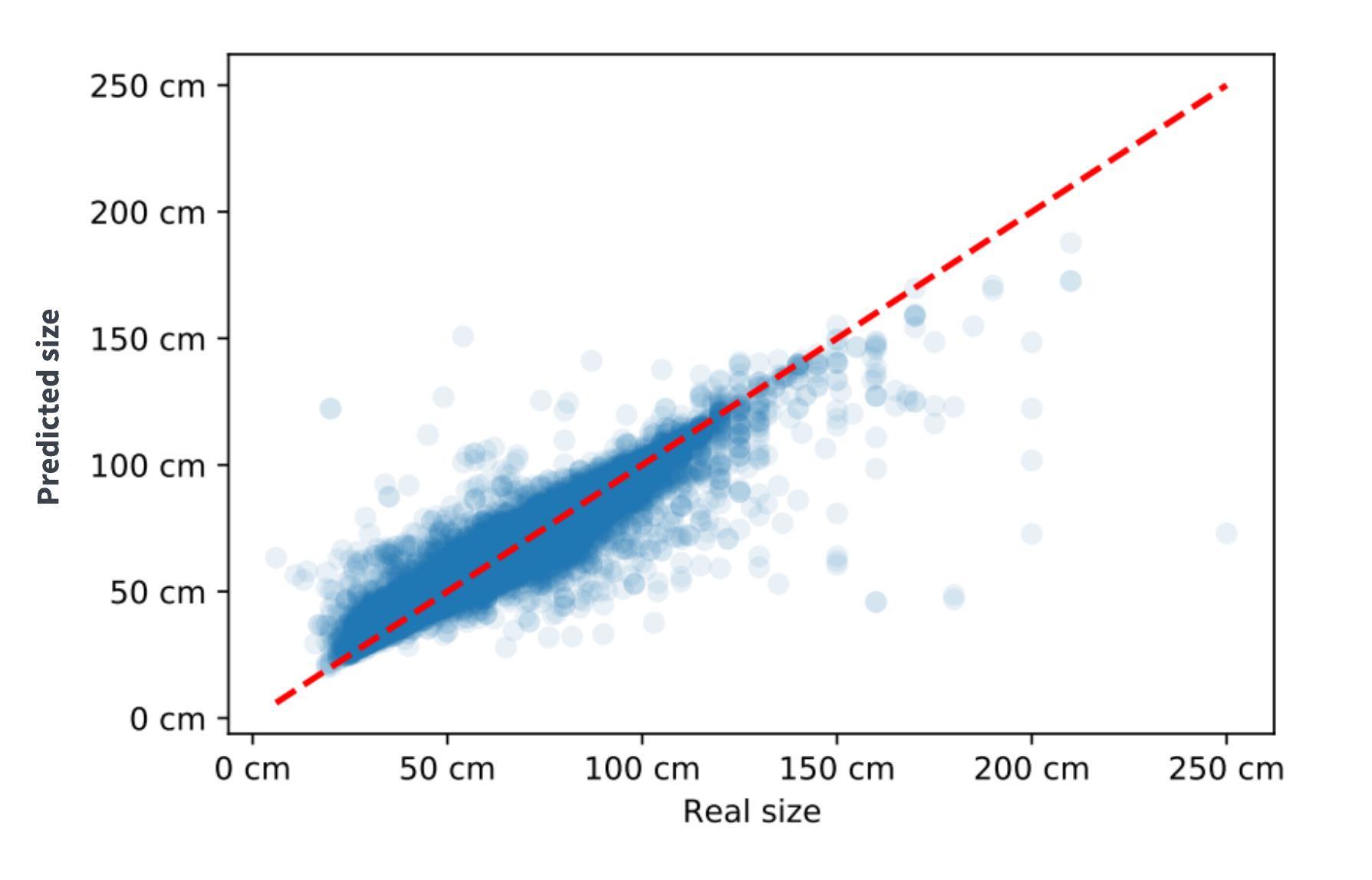}
\caption{Scatter plot of length predictions from the random forest vs ground truth (human annotator). Perfect predictions fall on the line $y=x$.}
\label{fig:real_pred}
\end{figure}

\begin{figure}
\includegraphics[width=\linewidth]{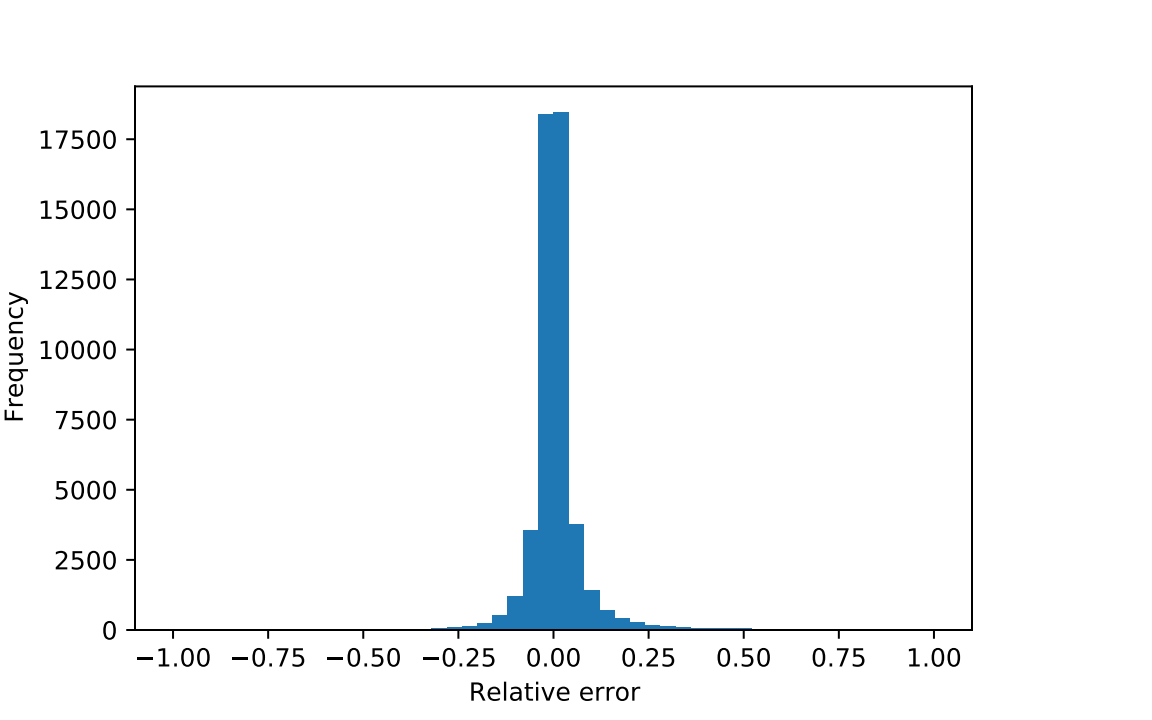}
\caption{Relative errors of the random forest model that predicts fish length.}
\label{fig:relative_errors}
\end{figure}

\begin{figure}
\includegraphics[width=\linewidth]{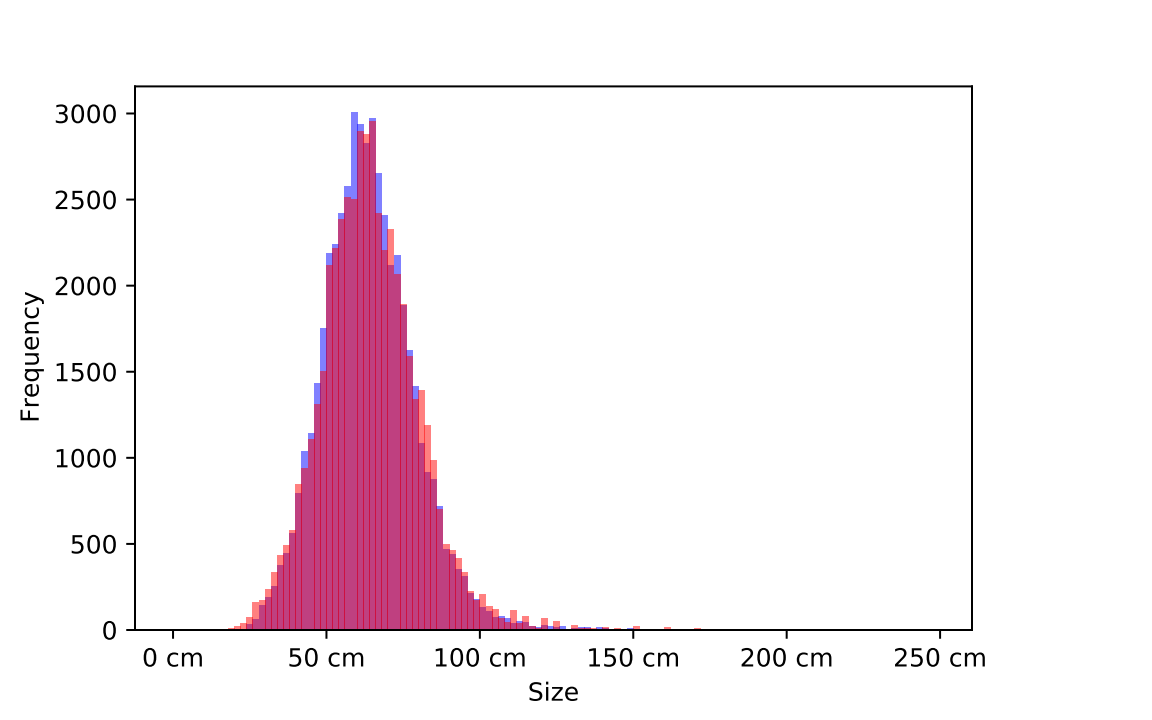}
\caption{The empirical distribution of predicted fish length (red) closely matches the data distribution (blue).}
\label{fig:dist_real_pred}
\end{figure}

Length regression was evaluated using five-fold cross-validation on the set of 50,000 single-fish images. The mean absolute error (MAE) was 2.3~cm and coefficient of determination ($R^{2}$) was 79\%. Figure~\ref{fig:real_pred} shows a scatter plot of predicted vs. true fish lengths on the validation set. The length estimates for mid-sized fish (60-80~cm), which are the majority in the validation set as shown in the distribution of size in Figure~\ref{fig:dist_real_pred}, are highly accurate. The model tends to underestimate larger fish sizes, likely due to a lack of large fish in the training data. This issue can be mitigated by including more large fish in the training set. Figures~\ref{fig:relative_errors} shows that the relative errors are symmetrically distributed around zero. This is critical for accurate length-based fish stock assessment methods, which evaluate the proportion of a population above or below certain length-based thresholds (such as size-at-maturity). We verify that the empirical distribution of the predicted fish lengths is close to the ground truth observations in Figure~\ref{fig:dist_real_pred}.

\section{Discussion and Future Research}\label{future_work}

The use of computer vision models for fish stock assessment could have great economic benefits by reducing time, cost, and labor. This paper presents FishNet: a system for performing fish counting, species classification, and size estimation from photographs. The system was developed and tested using a large collection of 300,000 annotated photographs generously provided by Yayasan Konservasi Alam Nusantara. While further work is needed to curate these annotated photographs to fully utilize them for machine learning, our results show that even training on a subset of this data yields promising results. One avenue for future research involves considering machine learning in the presence of label noise, as well as deep active learning methods for image classification \cite{motsoehli2023deepActiveLearningNoisy}, and segmentation under a noisy labeling oracle \cite{Han:CoTeachingNoisy18,Chicheng:ActiveStrongWeakLabelers15}. 
   
\section*{Acknowledgement}
We acknowledge the David and Lucile Packard Foundation for supporting this experiment and the CODRS data collection program (run by Yayasan Konservasi Alam Nusantara). The CODRS program has also received support from the Walton Family Foundation and the USAID Indonesia SNAPPER program (Cooperative Agreement No. AID\-497\-A\-16\-00011). We wish to thank technicians from People and Nature Consulting International for their help with the collection and labeling of the images and the group of over 500 Indonesian fishers who took pictures of their catch during their work at sea. We acknowledge technical support and computing resources from University of Hawaii Information Technology Services – Cyberinfrastructure funded in part by the National Science Foundation CC* award \texttt{\#}2201428.


\balance
\bibliographystyle{unsrt}
\bibliography{mybibliography}  

\end{document}